\documentclass[10pt,twocolumn,letterpaper]{article}

\usepackage[pagenumbers]{wacv} % To force page numbers, e.g. for an arXiv version

\usepackage{graphicx}
\usepackage{amsmath}
\usepackage{amssymb}
\usepackage{booktabs}

\usepackage{lipsum}
\usepackage{xcolor}
\usepackage{pifont}
\usepackage{array,multirow}
\newcolumntype{x}[1]{>{\centering\arraybackslash\hspace{0pt}}p{#1}}
\usepackage{colortbl}      % Additional table coloring options

\usepackage[accsupp]{axessibility}
\usepackage[pagebackref,breaklinks,colorlinks]{hyperref}

\makeatletter
\renewcommand*{\@fnsymbol}[1]{\ensuremath{\ifcase#1\or *\or \dagger\or \ddagger\or
   \mathsection\or \mathparagraph\or \|\or **\or \dagger\dagger
   \or \ddagger\ddagger \else\@ctrerr\fi}}
\makeatother

\usepackage[capitalize]{cleveref}
\crefname{section}{Sec.}{Secs.}
\Crefname{section}{Section}{Sections}
\Crefname{table}{Table}{Tables}
\crefname{table}{Tab.}{Tabs.}

\def\wacvPaperID{381} % *** Enter the WACV Paper ID here
\def\confName{WACV}
\def\confYear{2025}

\begin{document}

\title{Fine-Grained Spatial and Verbal Losses for 3D Visual Grounding}

\author{Sombit~Dey$^{1,2}$ \hspace{10px} Ozan~Unal$^{1,3}$\thanks{Corresponding author: Ozan Unal, ozan.unal@vision.ee.ethz.ch}
 \hspace{10px}
Christos~Sakaridis$^1$ \hspace{10px} Luc~Van~Gool$^{1,2,4}$
\\
$^1$ETH Zurich, $^2$INSAIT, $^3$Huawei Technologies, $^3$KU Leuven \\
}

\maketitle

%%%%%%%%% ABSTRACT
\begin{abstract}
3D visual grounding consists of identifying the instance in a 3D scene which is referred to by an accompanying language description. While several architectures have been proposed within the commonly employed grounding-by-selection framework, the utilized losses are comparatively under-explored. In particular, most methods rely on a basic supervised cross-entropy loss on the predicted distribution over candidate instances, which fails to model both spatial relations between instances and the internal fine-grained word-level structure of the verbal referral. Sparse attempts to additionally supervise verbal embeddings globally by learning the class of the referred instance from the description or employing verbo-visual contrast to better separate instance embeddings do not fundamentally lift the aforementioned limitations. Responding to these shortcomings, we introduce two novel losses for 3D visual grounding: a visual-level offset loss on regressed vector offsets from each instance to the ground-truth referred instance and a language-related span loss on predictions for the word-level span of the referred instance in the description. In addition, we equip the verbo-visual fusion module of our new 3D visual grounding architecture AsphaltNet with a top-down bidirectional attentive fusion block, which enables the supervisory signals from our two losses to propagate to the respective converse branches of the network and thus aid the latter to learn context-aware instance embeddings and grounding-aware verbal embeddings. AsphaltNet proposes novel auxiliary losses to aid 3D visual grounding with competitive results compared to the state-of-the-art on the ReferIt3D benchmark.
% AsphaltNet ranks first on the standard 3D grounding benchmarks of Nr3D and Sr3D.
\end{abstract}
\begin{figure}
    \centering
    \includegraphics[width=\columnwidth]{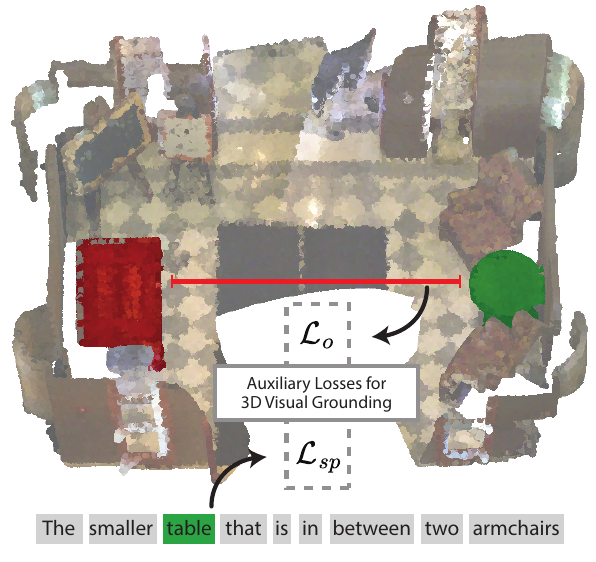}
    \caption{We propose two auxiliary losses to aid the task of 3D visual grounding: i) offset loss $\mathcal{L}_o$ between instances to promote better localization and improve object separability; ii) span loss $\mathcal{L}_{sp}$ on the prompts to provide a higher granularity of supervision.}
    \label{fig:teaser}
\vspace{-10px}\end{figure}
%%%%%%%%% BODY TEXT
% reduce couple of parahs
\section{Introduction}
\label{sec:intro}
With the ever-increasing popularity of the interaction- and 3D-heavy applications of augmented/virtual reality and human-support-oriented robotics in indoor home environments, bridging the gap between language understanding and 3D visual perception is becoming essential. This need has motivated a large body of recent works on the task of 3D visual grounding~\cite{zhao2021_3DVG_Transformer,yuan2021instancerefer,jain2022bottom,huang2021tgnn,roh2022languagerefer}, surveys papers ~\cite{liu2024survey}, which consists in identifying and localizing the object in the input 3D scene which is referred by a language description.

3D visual grounding works are classified into two main categories: (i) referral-based 3D object identification methods and (ii) referral-based 3D object localization methods. The former variant requires as visual inputs both the raw 3D point cloud and the ground-truth 3D bounding boxes or instance masks of all objects in the input scene which belong to the class of the referred object~\cite{achlioptas2020referit_3d}. On the contrary, the latter variant consists in the 3D localization of the referred object directly from the raw 3D point cloud~\cite{chen2020scanrefer}.
While the two variants might appear fundamentally different at first sight, referral-based localization reduces to referral-based identification after candidate object boxes or masks have been generated. 
3D grounding is typically performed in both cases in a grounding-by-selection fashion, using a verbo-visual fusion module which aims to classify the referred object based on the language description.

The de-facto choice in the literature~\cite{zhao2021_3DVG_Transformer,unal2023ways,yuan2021instancerefer,jain2022bottom,chen2020scanrefer} for supervising the above verbo-visual fusion module is via a supervised cross-entropy loss on the module's softmax classification outputs over the set of candidate objects. The limitations of this loss are that (i) it treats the candidate 3D objects as isolated entities that independently correlate with the language description, even though these objects co-exist in the 3D scene and participate in mutual spatial arrangements and relative configurations, and (ii) it views the language description as a monolithic entity, neglecting its internal structure. Some works~\cite{huang2022multiview,zhao2021_3DVG_Transformer,achlioptas2020referit_3d} extend supervision via an additional, language-driven cross-entropy loss on a class distribution of the referred object predicted from a global language embedding. While this language-driven loss can increase the awareness of language features about the semantics of the referred object, it does not lift limitation (ii) above, as it still operates on a global language embedding, which obscures the attribution of gradients to individual word embeddings that are more closely associated with the referred object. Another line of works~\cite{jain2022bottom,unal2023ways} proposes contrastive losses between verbal and visual embeddings to inject a better structure in the joint verbo-visual embedding space. However, such cross-modal contrast improves separation between distinct instance embeddings only \emph{indirectly} by pulling them to / pushing them away from sentence~\cite{unal2023ways} or word~\cite{jain2022bottom} embeddings and it still treats each instance individually and not jointly with others, leaving limitation (i) above unaddressed.
% consisting of multiple words, especially noun phrases, which can in principle refer to different 3D objects and their mutual relations, including the actual object of interest that the entire utterance refers to. 
% This selection step is commonly supervised via a cross-entropy loss that penalizes the network for its confidence on non-target objects. We argue that while such a loss shows promising results in the literature, it does not explicitly promote better utilization of language or aid spatial localization.

In this work, we attempt to smooth the loss manifold for 3D visual grounding by proposing two novel losses to overcome the two aforementioned limitations of the basic supervised grounding-by-selection setup. First, we propose to make 3D instances aware of spatial visual grounding context by letting each of them regress a 3D offset vector pointing to the ground-truth referred instance. In this way, we encourage the instance embeddings to attend to each other in a way that is informed about the \emph{spatial} location to which the semantics of the referral is grounded. The specific implementation we follow for the verbo-visual fusion module involves a basic attention-based building block. As we aim at injecting the aforementioned grounding awareness into embeddings from an early-on stage of the verbo-visual fusion module, we add offset vector regression heads after each such attention block and supervise each of them with a distinct offset regression loss. For the attention-based block, we build on the generic design proposed in~\cite{unal2023ways} but reverse the bottom-up masked attention into a top-down counterpart. In this transformer architecture, earlier attention blocks are allowed to attend to a larger spatial context of instances in order to be able to already predict potentially large offsets if necessary, while later blocks are confined to a narrower attention mask, as they have already attended to more distant parts of the scene in earlier stages.

% To tackle referral-based 3D object identification, we build on top of Unal~\etal~\cite{unal2023ways} to establish a strong baseline that we call AsphaltNet. To improve the training process of AsphaltNet, we propose appending a regression head that for each object in a scene predicts their offset to the referred target instance. An offset loss $\mathcal{L}_{o}$ can then be employed to supervise the network. With such an auxiliary target, we aim to promote grounding-aware attention between instances within the transformer-based fusion module.

Second, we propose to exploit the input language description at a finer level of granularity than the previously considered global sentence level by supervising the predicted word-level \emph{span} of the ground-truth referred object in the description, which is produced by the final verbal embeddings of the verbo-visual fusion module. These spans are normally available in standard 3D visual grounding datasets~\cite{achlioptas2020referit_3d}, but they have been largely overlooked by virtually all existing works when designing their models. By applying our proposed span loss, we can precisely attribute the gradient of each of the terms of the loss back to the embedding of the corresponding word, treating words inside the span differently from words outside it. This makes our word embeddings aware of the verbally described characteristics of the referred object. To allow this verbal awareness to be combined with visual awareness and thus produce word embeddings that are discriminative for the referred instance, we additionally enable cross-attention in the direction from the visual branch to the verbal branch in the attention-based building blocks of our verbo-visual fusion module, in a bidirectional attention formulation similar to that in~\cite{jain2022bottom}. Put together, our two novel losses along with top-down bidirectional attentive fusion form our complete AsphaltNet architecture for 3D visual grounding. AsphaltNet is evaluated on the Nr3D and Sr3D benchmarks and compared to the state-of-the-art approaches.
% AsphaltNet ranks $1^{st}$ on the Nr3D and Sr3D benchmarks.

\section{Related Work}
%%% get rid of 2 parahs from here, 
\label{sec:related_work}
\par{3D visual grounding} is one of the core problems that lies in the intersection between computer vision and language, aiming to localize objects in 3D space that are referred to by natural language prompts. To date, its 2D counterpart has been extensively studied~\cite{hu2016natural,2d1,mao2016generation, plummer2015flickr30k}. In the literature, 3D visual grounding is explored in two ways: (i) by purely focusing on the verbo-visual fusion, where given all available ground truth objects a selection task is carried out to identify the best fitting candidate, i.e. referral-based object identification~\cite{achlioptas2020referit_3d}; or (ii) in an end-to-end fashion, where given a language utterance, a referred object is localized within the 3D scene via a bounding box or instance mask, i.e. referral-based object localization~\cite{chen2020scanrefer}. While seemingly different, a common strategy when tackling referral-based object localization is to employ a grounding-by-selection approach, where after a visual backbone produces candidate objects, the task boils down to an identification problem~\cite{chen2020scanrefer, zhao2021_3DVG_Transformer, unal2023ways}. In this work, we undertake referral-based object identification to focus on improving the training process for multi-modal training.

The seminal works of Chen~\etal~\cite{chen2020scanrefer} and Achlioptas~\etal~\ref{tab:results_referit} provided the first 3D visual grounding benchmarks of ScanRefer, Nr3D and Sr3D by providing grounding annotations for the popular indoor RGB-D dataset ScanNet~\cite{dai2017scannet}. In ScanRefer, a baseline model was introduced that tackled the fusion of language and vision in a rudimentary way, where each modality is individually encoded to later be joined via concatenation and processed via a shallow MLP. 
% 3DVG-Transformer~\cite{zhao2021_3DVG_Transformer} built on top of the proposed architecture, introducing a transformer based fusion module that relied on cross-attention to fuse the two streams of information. Furthermore 3DVG-Transformer modelled the proximity relations between instances and fused the spatial proximity information with the attention matrix for improved grounding performance. 

ConcreteNet~\cite{unal2023ways} improved 3DVG-Transformer by (i) by utilizing bottom-up attentive fusion module that localized mask attention in a progressively increasing neighborhood, (ii) by implicitly promoting separation between instance candidates via a cross-modal contrastive loss, and (iii) by tackling view-dependent prompts via a novel learned global camera token. In our work, we utilize a similar mask attention block that progressively scales in size, but do so in a top-down manner. Furthermore, instead of a multiplex attention layer, we employ bidirectional cross-attention to not only pass information from the natural language description to the 3D scene, but vice versa.

3D grounding solely from 2D image inputs was addressed in~\cite{prabhudesai2020embodied} which a scene-graph based 3D grounding was explored in~\cite{feng2021free}. Joint multi-task architectures for 3D dense captioning and 3D visual grounding were presented in~\cite{cai20223djcg}, in~\cite{chen2022d3net}, While most 3D grounding approaches decouple object proposal generation from verbo-visual fusion, \cite{luo20223d} proposes a single-stage approach which does not involve vision-only prediction of object proposals but performs mid-level fusion of language with 3D visual features.

InstanceRefer~\cite{yuan2021instancerefer} constructs attention based verbo-visual fusion between global sentence embedding and instance embeddings, and tackled dense 3D visual grounding, i.e. referral-based 3D instance segmentation. TransRefer3D~\cite{He_2021} proposes an entity-aware attention module and a relation-aware attention module to conduct fine-grained cross-modal feature matching. 
LanguageRefer~\cite{roh2022languagerefer} uses combines spatial embeddings extracted from provided ground truth bounding boxes with fine-tuned language embeddings from DistilBert~\cite{sanh2019distilbert} to identify the referred object. Furthermore, they analyze view-dependent utterances and viewpoint annotations.
SAT~\cite{yang2021sat} utilizes 2D image semantics in the training stage to ease verbo-visual joint representation learning and assist 3D visual grounding. LAR~\cite{bakr2022look} tackles a similar problem of assisting 3D feature extraction by incorporating 2D object information. However in contrast to SAT, LAR does not rely on the existence of available 2D data, but synthetically generates 2D clues from 3D point clouds. MVT~\cite{huang2022multiview} repeatedly augments the 3D scene to generate a multiple view input and simultaneously processes and then aggregates the results. MDETR~\cite{kamath2021mdetr} is a language grounding model for 2D images that decodes object boxes and aligns them with relevant spans in input utterances. BUTD-DETR~\cite{jain2022bottom} builds upon MDETR but relies on a 3D object detector to generate referential grounding results, which are then grounded in the visual scene, augmenting supervision for the referential grounding.
% BUTD-DETR utilizes a bidirectional attention layer within its architecture, allowing information flow between the visual and verbal inputs. In our work, similar to BUTD-DETR we employ bidirectional cross-attention layers to enable communication from the 3D scene to the verbal tokens to more effectively utilize the span loss. Unlike BUTD-DETR, we employ a top-down spherical masking operation within our framework to localize the attention and improve grounding performance when dealing with referrals that distinguish objects via neighboring relations.
BUTD-DETR [20], builds upon  MDETR~\cite{kamath2021mdetr}, utilizing a 3D object detector for referential grounding, leveraging bidirectional attention layers to integrate visual and verbal inputs. Similar to BUTD-DETR, our work employs bidirectional cross-attention to allow information flow between visual tokens and verbal tokens. However, we introduce a top-down spherical masking technique to localize attention, improving grounding performance in referrals that require distinguishing objects based on neighboring relations.
ScanEnts3D~\cite{abdelreheem2024scanents3d} advances the field of referential grounding by providing ground truth anchors, enabling the study of the effects of anchors and distractors for 3D grounding.
3DVista~\cite{3dvista} aims to tackle 3D visual grounding by utilizing aggregation of datasets and tasks with a unified pre-trained transformer for 3D visual and text alignment.
\begin{figure*}[t]
    \centering
    \includegraphics[width=.9\textwidth]{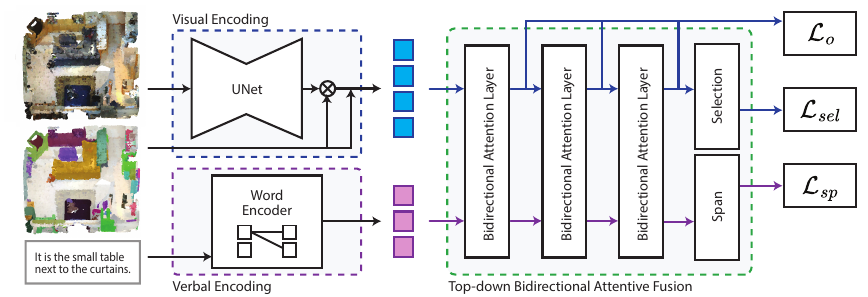}
    \caption{Illustration of the AsphaltNet pipeline. Via a UNet, we first compute per point features from the 3D point cloud and construct instance tokens via mean pooling on the provided instance masks and concatenation with the corresponding bounding box centroid and mean color vector. We encode the verbal input and generate word tokens using a pretrained LLM. A top-down bidirectional attentive fusion module consumes the instance and word tokens to ground the natural language description within the scene, i.e. predict the referred object. To aid the training process, we employ two auxiliary losses: $\mathcal{L}_o$ that aids localization, and $\mathcal{L}_{sp}$ to supervise the language encoding.}
    \label{fig:pipeline}
\vspace{-10px}\end{figure*}
\section{Method}

In this section we introduce our model AsphaltNet. First, we individually encode the visual and verbal cues (Sec.\ref{subsection:Vi} and Sec.\ref{subsection:Ver} respectively), which form the input for the verbo-visual fusion module that processes the two modalities to select the referred instance (Sec.~\ref{sec:fusion}). Later we explore two auxiliary losses to aid the training of the model (Sec~\ref{sec:aux_loss}). An illustration can be found in Fig.~\ref{fig:pipeline}.

\subsection{Encoding Visual Cues}
\label{subsection:Vi}
Typically, segmentation models operate directly on available dense information, which can be effectively encoded to extract high-fidelity scene-level features and decoded to generate dense predictions~\cite{cvpr2017pointnet, qi2017pointnet++, wacv2021improving, iccv2019kpconv}. However, transformer-based verbo-visual fusion modules are heavily constrained by the number of the input tokens due to the considerable memory and computing requirements of self- and cross-attention operations~\cite{vaswani2017attention, dosovitskiy2020image}. To this end, a common solution is to construct a sparse representation of the scene that can capture relevant information for visual grounding~\cite{unal2023ways, zhao2021_3DVG_Transformer, jain2022bottom}.

In referral-based 3D object identification, the visual input is given by a point cloud $P \in \mathbb{R}^{N\times6}$ with $N$ points, each described by its 3D coordinates $(x,y,z)$ and color values $(r,g,b)$, as well as the ground truth instance masks $I \in \{0,1\}^{K\times N}$ with $K$ instances. Following set precedent~\cite{unal2023ways, zhao2021_3DVG_Transformer, jain2022bottom}, to avoid directly operating on the large point cloud data during verbo-visual fusion, we construct instance embeddings using the available mask information to act as visual tokens. Specifically, we extract high-level features from the input point cloud via a UNet backbone and mean pool pointwise features for each instance mask.
To preserve spatial and color information, we concatenate the center coordinates of the axis aligned bounding box along with the mean color vector with the pooled UNet features.

Formally, the instance embeddings $e_i$ that form the visual input for the verbo-visual fusion are given by:
\begin{equation}
\begin{split}    
    e_i &= [f_i,c,a] \in \mathbb{R}^{d+6} \textrm{, \ with} \\
    f_i &= (I \times \phi_{UNet}(P)) \oslash (I \times [1,1,1,..]^T)
\end{split}
\end{equation}
$c = (c_x,c_y,c_z)$ and  $a = (r,g,b)$ denoting the center and mean color of the instance candidate respectively, $\phi_{UNet}$ denoting the UNet encoding the produces a $d$ dimensional output and $\oslash$ the Hadamard division.

\subsection{Encoding Verbal Cues}
\label{subsection:Ver}

Large pretrained transformers have risen in popularity in recent years following their wide success in natural language processing~\cite{vaswani2017attention, liu2019roberta, song2020mpnet}. This success has carried over to computer vision applications that require or benefit from verbal cues. Given that such models are trained with large-scale datasets they are capable of capturing context and intent within natural language prompts. While most early work in 3D visual grounding utilize a more traditional approach in text encoding, namely GloVE~\cite{pennington2014glove}, recent efforts have moved on to transformer-based encoders such as BERT~\cite{devlin2018bert}, RoBERTa~\cite{liu2019roberta} or MPNet~\cite{song2020mpnet}.

We employ the pretrained BERT to initially encode word tokens. The  features are then projected using a $1\times1$ convolutional layer to match the dimension of the instance candidates $e_w \in \mathbb{R}^{|W| \times (d+6)}$ which form the verbal input tokens of our fusion module.

\subsection{Verbo-visual Fusion} \label{sec:fusion}

In referral-based 3D object localization, verbal and visual features are processed, to output a likelihood vector over the input candidates to ground the natural language prompt. In the current literature, a common strategy to fuse the two modalities is through the employment of attention:
\begin{equation} \label{eq:attention}
    g = \textrm{softmax}(q k^T) v
\end{equation}
with residual attention features $g$, the queries $q$ extracted from instance embeddings $e_i$ and key-value pairs $k$ and $v$ extracted from word features $e_w$ for cross-attention or from instance features $e_i$ for self-attention. The cross-attention allows information extracted from the pretrained verbal model to be routed to each  instance candidate. Subsequent self-attention layers that are formed between instance candidates routes this information, allowing the tokens to extract further relevant features for visual grounding.

\vspace{5px}
\noindent \textbf{Top-down Bidirectional Attentive Fusion (TBA):} Often natural language prompts localize a repetitive object, i.e. an object that is semantically not unique within a scene, by establishing its relation
to another object. Such relational cues are commonly constrained within local neighborhoods, e.g. ``Chair next to the door", or ``The ottoman in front of the sofa". While second nature to us, such localized information routing is difficult to learn for transformer-based architectures that rely on global information routing schemes as found in self-attention layers. To combat this, inspired by Mask2former~\cite{cheng2022masked}, ConcreteNet~\cite{unal2023ways} proposes using masked self-attention layers to explicitly limit information routing within local neighborhoods. Formally, Eq.~\ref{eq:attention} is restated as:
\begin{equation}
\begin{split}
        g &= \textrm{softmax}(M + q k^T) v \textrm{, \ with} \\
        M(i,j) & = \begin{cases}
        0, & \textrm{if } || c_i - c_j || < r \\
        -\infty,\quad & \textrm{otherwise}
        \end{cases}
\end{split}
\end{equation}
with $r$ denoting the radius of the spherical masking set per layer. Similarly, we construct a transformer-based verbo-visual fusion module by iteratively stacking masked self-attention layers followed by cross attention layers.

\begin{figure}[t]
    \centering
    \includegraphics[width=\columnwidth]{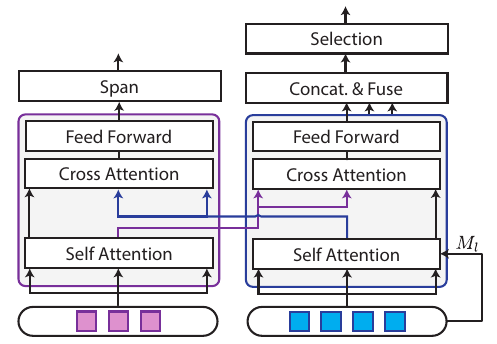}
    \caption{Illustration of top-down bidirectional attentive fusion.}
    \label{fig:tba}
\vspace{-10px}\end{figure}

Yet, our fusion module differs from that of ConcreteNet~\cite{unal2023ways} in two key aspects: (i) we utilize bidirectional cross-attention, i.e. cross-attention not only to route information from verbal cues to the visual, but also to allow the verbal cues to learn about the 3D space. We argue that while the features extracted from the pretrained verbal model are rich in information, the instance features that describe the 3D scene can provide strong priors to the word tokens and help to resolve view-dependent relations. Bidirectional cross-attention has previously shown to strongly aid 3D grounding performance~\cite{jain2022bottom} but was never utilized in a local attention setting; (ii) next, we employ a top-down approach when masking compared to a bottom-up approach, i.e. after each bidirectional attention block, starting from global attention we reduce the radius $r$ to further limit the spherical neighborhood. This is later discussed in Sec.~\ref{sec:aux_loss} - Offset Loss. An illustration of TBA can be seen in Fig.~\ref{fig:tba}.

\vspace{5px}
\noindent \textbf{Selection Loss:} Following the TBA, a $1\times1$ convolutional layer maps each embedding $e_i$ to a confidence score in order to form the prediction $u$. The prediction is supervised by the ground truth target $\hat{u}$ via a cross-entropy loss:
\begin{equation}
 \mathcal{L}_{sel} = H(u, \hat{u}).
\end{equation}

\subsection{Auxiliary Losses for 3D Visual Grounding}
\label{sec:aux_loss}

\begin{figure*}[t]
    \centering
    \includegraphics[width=.9\textwidth]{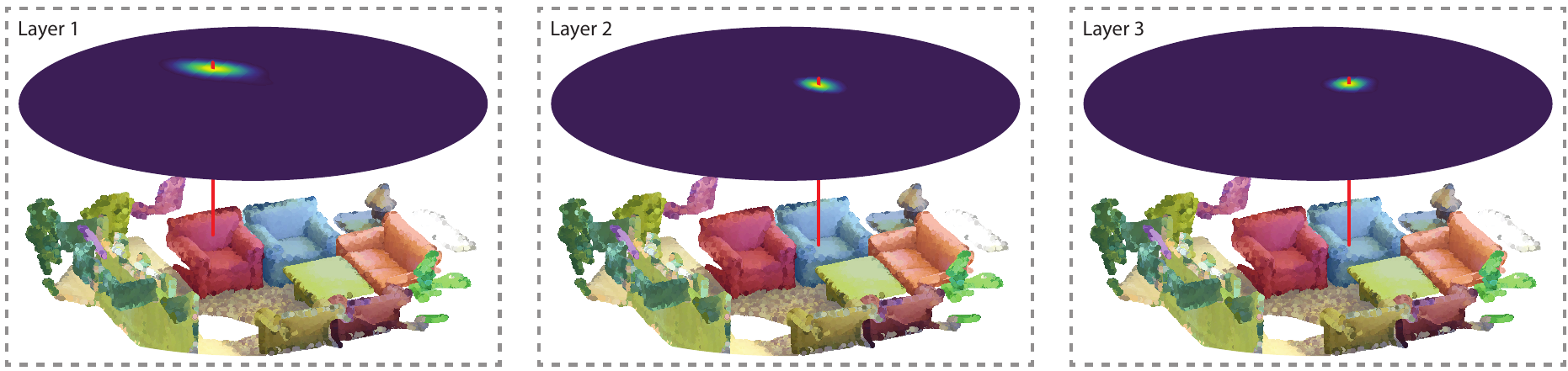}
    \caption{Sample offset predictions of AsphaltNet from the Nr3D~\cite{achlioptas2020referit_3d} \textit{val}-set given the prompt ``Facing the armchairs, the chair in the upper left". We overlay the heatmap of the predicted offset values per instance to the target on top of the scene for all the three layers of the fusion module. The offset prediction is refined at each step as the distribution of predicted offsets tightens and converges on the target instance.}
    \label{fig:offset_loss}
\end{figure*}

The literature is currently focused on exploring new architectural setups to better ground natural language prompts to 3D scenes~\cite{jain2022bottom, yang2021sat, zhao2021_3DVG_Transformer}. The respective training objectives remain highly under-explored. Similarly as Sec.~\ref{sec:fusion}, current methods are supervised only via a limited pool of losses such as (i) cross-entropy that aims to either maximize the softmax probability of the ground-truth object when applied on the instance tokens, or (ii) identify the semantic class of the target object when applied on the word tokens; and (iii) contrastive loss that aims to map matching word and instance tokens to close representations. We argue that while these losses have shown promising results, they remain insufficient as they do not explicitly promote localization of the prompt within the 3D scene or simply remain too coarse for informative supervision.

\vspace{5px}
\noindent\textbf{Offset Loss ($\mathcal{L}_o$):} To remedy this, inspired by common practices in 3D instance segmentation~\cite{ngo2023isbnet,wu20223d}, we propose to define an offset loss at the level of objects, where each object in the scene is required to predict its offset to the referred object, in order to better ground the natural language in 3D space. Formally, we define the offset loss as:
\begin{equation}
    \mathcal{L}_o = \sum_{i=1}^{|I|} \mathrm{L}_2(o_i, (c_i - c_{gt}))
\end{equation}
with $o_i$, $c_i$ denoting the predicted offset value and centroid for instance $i$ and $c_{gt}$ the referred object centroid.

In essence, in order for each instance to be able to predict the offset to the referred object, grounding-aware features are learned for all instances in the self-attention layers. With such an auxiliary loss, we not only achieve better localization through direct supervision, but also implicitly promote separation between the features of the referred object and those of spatially distant repetitive objects which may bear semantic similarity to the former.

We employ the offset loss after every bidirectional attentive fusion block to encourage the network the extract grounding-aware instance features at varying spherical neighborhoods. It is for this reason that a top-down approach is required, as in a bottom-up setting, where masking is carried out in a local-then-global fashion, early layers do not allow information routing across the entire scene, which limits the ability of an instance candidate to predict an accurate offset to a distant referred object. In Fig.~\ref{fig:offset_loss} we showcase example offset predictions from AsphaltNet.

\vspace{5px}
\noindent\textbf{Span Loss ($\mathcal{L}_{sp}$):} To better exploit the input verbal utterances at a finer level of granularity than the aforementioned considerations that supervise the model at a sentence level through cross-entropy, we propose predicting and supervising on a word-level through the \textit{span} of a description. The span $\hat{l} \in \{0,1\}^{|W|}$ is a vector with entry 1 for where the word or words describe the referred object, and 0 otherwise. Formally we define the span loss as:
\begin{equation}
    \mathcal{L}_{sp} = H^1(l,\hat{l})
\end{equation}
with $H^1$ denoting the binary cross entropy loss, $l$ the ground truth span labels and $\hat{l}$ the predicted span logits.

The span loss allows precisely gradient flow from each loss term back to the corresponding word embedding. Each words inside the span is treated differently from words outside it, increasing awareness towards the verbally described characteristics of the referred object.

\begin{table*}[t]
\centering
  \tabcolsep=0.26cm
\begin{tabular}{|c|l|c|x{1.2cm}|x{1.3cm} x{1.1cm} x{1.1cm} x{1.3cm}|}
\cline{2-8}
\multicolumn{1}{c|}{} & Method & Input &  \textbf{Overall} & Easy & Hard & V-Dep. & V-Ind. \\
\hline
\multirow{12}{*}{\rotatebox[origin=c]{90}{Nr3D}} & ReferIt3D~\cite{achlioptas2020referit_3d} & 3D & 35.6 & 43.6 & 27.9 & 32.5 & 37.1 \\
% ReferIt3D~\cite{achlioptas2020referit_3d} &  3D & 35.6 & 43.6 & 27.9 & 32.5 & 37.1 \\
% & Text-Guided-GNNs~\cite{} & 37.3 & 44.2 & 30.6 & 35.8 & 38.0 \\
% & InstanceRefer~\cite{yuan2021instancerefer}  & 3D&  38.8 & 46.0 & 31.8 & 34.5 & 41.9 \\
% & 3DRefTransformer~\cite{} & 39.0 & 46.4 & 32.0 & 34.7 & 41.2 \\
& 3DVG-Transformer~\cite{zhao2021_3DVG_Transformer} & 3D&  40.8 & 48.5 & 34.8 & 34.8 & 43.7 \\
% & FFL-3DOG~\cite{} & 41.7 & 48.2 & 35.0 & 37.1 & 44.7 \\
% & TransRefer3D~\cite{He_2021}  &3D &  42.1 & 48.5 & 36.0 & 36.5 & 44.9 \\
& LanguageRefer~\cite{roh2022languagerefer}  & 3D &  43.9 & 51.0 & 36.6 & 41.7 & 45.0 \\
% & LAR~\cite{bakr2022look} & 3D&  48.9 & 56.1 & 41.8 & 46.7 & 50.2 \\
& ConcreteNet~\cite{unal2023ways} & 3D & 45.5 & 52.0 & 38.7 & 38.9 & 48.3 \\
& SAT~\cite{yang2021sat} & 2D+3D&  49.2 & 56.3 & 42.4 & 46.9 & 50.4 \\
& BUTD-DETR~\cite{jain2022bottom}  & 3D&  54.6 & 60.7 & 48.4 & 46.0 & 58.0 \\
& MVT~\cite{huang2022multiview}  & MV-3D & 55.1 & 61.3 & 49.1 & 54.3 & 55.4 \\
  & 3DVista (scratch) & 3D & 57.5 & 65.9&  49.4 & 53.7 & 59.4 \\

\cline{2-8}
& AsphaltNet  & 3D & \textbf{58.9} & \textbf{64.3} & \textbf{53.9} & \textbf{57.4} & \textbf{59.8} \\
\cline{2-8}
& \textcolor{gray}{ ScanEnts~\cite{abdelreheem2024scanents3d}} & \textcolor{gray}{3D} & \textcolor{gray}{59.3} & \textcolor{gray}{65.4} & \textcolor{gray}{53.5} & \textcolor{gray}{57.3} & \textcolor{gray}{60.4} \\
  & \textcolor{gray}{3DVista~\cite{3dvista}} & \textcolor{gray}{3D*} &  \textcolor{gray}{64.2} & \textcolor{gray}{72.1} & \textcolor{gray}{56.7} & \textcolor{gray}{61.5} & \textcolor{gray}{65.1} \\
& \textcolor{gray}{Vil3DRel~\cite{chen2022language}} & \textcolor{gray}{3D} & \textcolor{gray}{64.4} & \textcolor{gray}{70.2} & \textcolor{gray}{57.4} & \textcolor{gray}{62.0} & \textcolor{gray}{64.5} \\

% utilizes llm based prompt tuning, 
\hline
\hline
\multirow{10}{*}{\rotatebox[origin=c]{90}{Sr3D}} & ReferIt3D~\cite{achlioptas2020referit_3d}  &3D & 40.8 & 44.7 & 31.5 & 39.2 & 40.8 \\
% & Text-Guided-GNNs~\cite{} & 45.0 & 48.5 & 36.9 & 45.8 & 45.0 \\
% & 3DRefTransformer~\cite{} & 47.0 & 50.7 & 38.3 & 44.3 & 47.1 \\
% & InstanceRefer~\cite{yuan2021instancerefer}  & 3D&  48.0 & 51.1 & 40.5 & 45.4 & 48.1 \\
& 3DVG-Transformer~\cite{zhao2021_3DVG_Transformer} &  3D &  51.4 & 54.2 & 44.9 & 44.6 & 51.7 \\
& LanguageRefer~\cite{roh2022languagerefer} & 3D &  56.0 & 58.9 & 49.3 & 49.2 & 56.3 \\
% & TransRefer3D~\cite{He_2021}  & 3D &  57.4 & 60.5 & 50.2 & 49.9 & 57.7 \\
& SAT~\cite{yang2021sat} & 2D+3D &  57.9 & 61.2 & 50.0 & 49.2 & 58.3 \\
% & LAR~\cite{bakr2022look} & 3D&  59.4 & 63.0 & 51.2 & 50.0 & 59.1 \\
& NS3D~\cite{hsu2023ns3d} & 3D& 62.7 & 64.0 & 59.6 & 62.0 & 62.7 \\
& MVT~\cite{huang2022multiview}& MV-3D &  64.5 & 66.9 & 58.8 & 58.4 & 64.7 \\
& BUTD-DETR~\cite{jain2022bottom}  &3D &  67.0 & 68.6 & 63.2 & 53.0 & 67.6 \\
  & 3DVista (scratch) & 3D & 69.6 & 72.1 & 63.6 & 57.9&  70.14 \\

\cline{2-8}
& AsphaltNet & 3D &  \textbf{69.7} & \textbf{71.9} & \textbf{64.5} & \textbf{67.2} & \textbf{70.0}\\
\cline{2-8}

& \textcolor{gray}{Vil3DRel~\cite{chen2022language}} & \textcolor{gray}{3D} & \textcolor{gray}{72.8} & \textcolor{gray}{74.9} & \textcolor{gray}{67.9} & 	\textcolor{gray}{63.8} &	\textcolor{gray}{73.2}\\
% & \textcolor{gray}{Cot3DRef~\cite{bakr2023cot3dref}}  & \textcolor{gray}{3D } & \textcolor{gray}{73.2} &	\textcolor{gray}{75.2} &	\textcolor{gray}{67.9}&	\textcolor{gray}{67.6}&	\textcolor{gray}{73.5} \\ 

& \textcolor{gray}{3DVista~\cite{3dvista}} & \textcolor{gray}{3D*} & \textcolor{gray}{76.4 } &	\textcolor{gray}{78.8}&	\textcolor{gray}{71.3}&	\textcolor{gray}{58.9}&	\textcolor{gray}{77.3}\\
\hline
\end{tabular}
\caption{3D visual grounding results for referral-based object identification on the Nr3D and Sr3D datasets~\cite{achlioptas2020referit_3d}. V-Dep. and V-Ind. refer to the view-dependent and independent subcategories. MV refers to multi-view. 3D* refers to the use of additional 3D scenes. ScanEnts~\cite{abdelreheem2024scanents3d} is evaluated along with GT anchor proposals. Vil3Dref has a reduced evaluation protocol, excluding challenging longer prompts.} 
\label{tab:results_referit}
\vspace{-10px}\end{table*}

\begin{figure*}
    \centering
    \includegraphics[width=.9\textwidth]{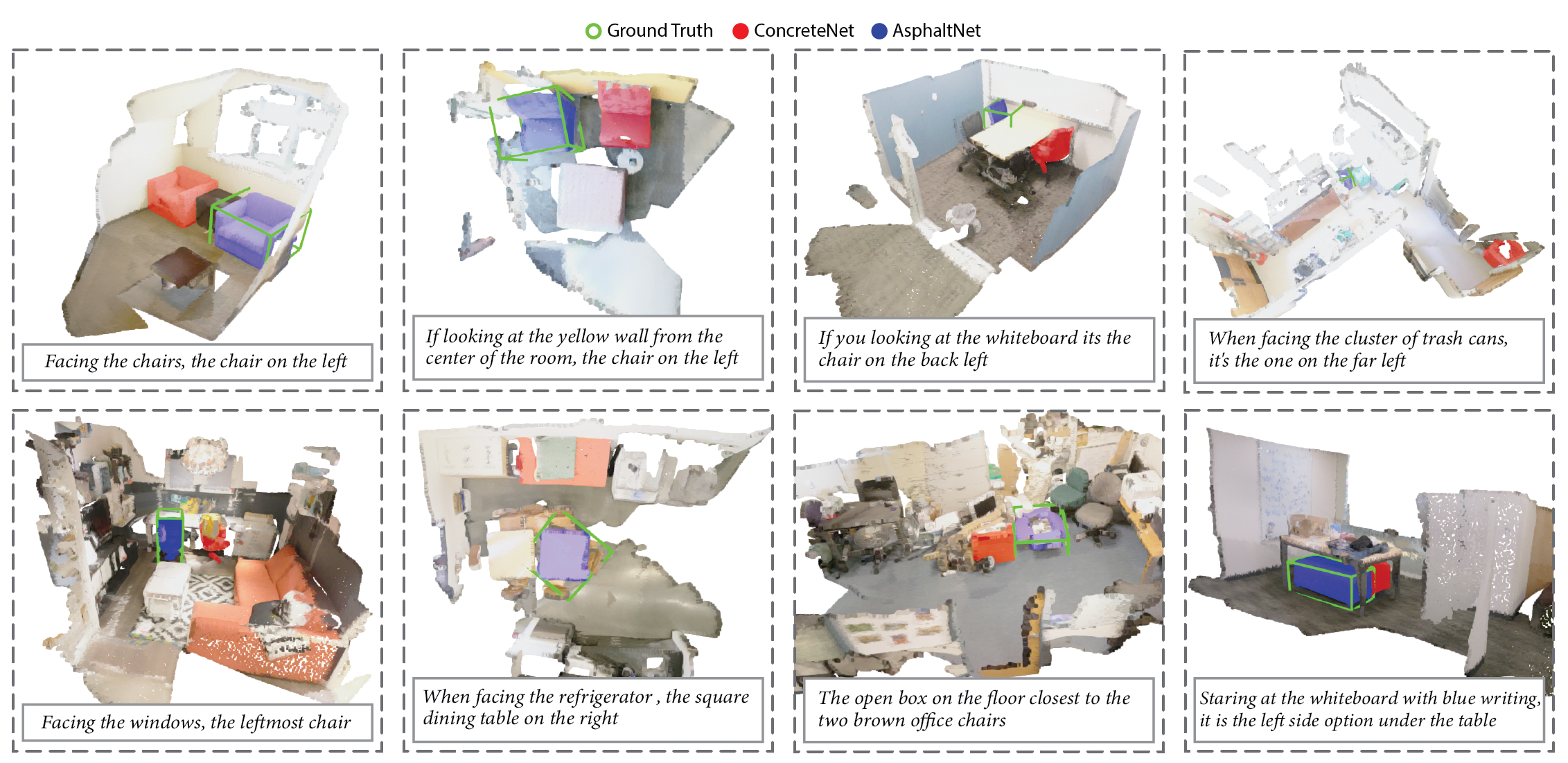}
    \caption{Qualitative results on the Nr3D \textit{val}-set comparing AsphaltNet with the baseline ConcreteNet model. We illustrate the predicted instances via instance masks (red) for ConcreteNet, and (blue) for AsphaltNet. In green, we provide the axis aligned bounding box of the referred ground truth object.}
    \label{fig:results}
\vspace{-10px}\end{figure*}

\section{Experiments}
\subsection{Implementation details}
% For the visual encoder, we use the 3D Unet backbone of
% ISBNet [7] pretrained on ScanNet and frozen.
% We employ standard augmentation strategies such as rotation, translation and cropping. For the verbal
% backbone we use a pretrained frozen BERT~\cite{devlin2018bert}. Following Zhao et al.~\cite{zhao2021_3DVG_Transformer}, we mask the referred object nouns with
% a probability of $0.2$. We also employ text augmentation by
% randomly replacing word with their synonyms with a probability of $0.3$. For the fusion module, we use a dimension
% of 128 across all layers, using 2 transformer encoder layers for the language abstraction with sinusoidial positional
% encoding. TBA consists of 3 bidirectional attention
% layers, each with a set spherical radius of [$\infty$, $2.5$m, $1$m]
% following the inverted scheme of ConcreteNet~\cite{unal2023ways}. We use
% a batch size of 1 with each batch consisting of up to 64 ut-
% terances. This allows us to process only a single scene at
% a time, reducing total memory requirements while retain-
% ing clean gradients for grounding. 
% We train for 300 epochs
% using AdamW with a learning rate of 0.0001 on a single
% Nvidia Geforce RTX 3090. 
% We provide the implementation details and dataset information in the supplement.

\noindent \textbf{Implementation details:} For the visual encoder, we use the 3D Unet backbone of ISBNet~\cite{ngo2023isbnet} pretrained on ScanNet and frozen. We employ standard augmentation strategies such as rotation, translation and cropping. For the verbal backbone we use a pretrained frozen MPNet~\cite{song2020mpnet}. Following Zhao~\etal~\cite{zhao2021_3DVG_Transformer}, we mask the referred object nouns with a probability of 0.2. We also employ text augmentation by randomly replacing word with their synonyms with a probability of 0.3~\cite{EDA}. For the fusion module, we use a dimension of 128 across all layers, using 2 transformer encoder layers for the language abstraction with sinusoidial positional encoding. TBA consists of 3 bidirectional attention layers, each with a set spherical radius of [$\infty, 2.5m, 1m$] following the inverted scheme of ConcreteNet~\cite{unal2023ways}. We use a batch size of 1 with each batch consisting of up to 64 utterances. This allows us to reduce the total memory requirements while retaining clean gradients for grounding. We train for 300 epochs using AdamW with a learning rate of 0.0001 on a single Nvidia Geforce RTX 3090.

% \noindent \textbf{Dataset:} We evaluate our method on the Nr3D and Sr3D datasets~\cite{achlioptas2020referit_3d} that provide human-annotated and synthetically generated utterances to identify instances on ScanNet~\cite{dai2017scannet}.
% Specifically, Nr3D consists of 41,503 natural language queries and Sr3D consists of 83,572 descriptions over 707 unique indoor scenes referring to 76 object classes. For evaluation, the dataset is split in two ways. First split considers the difficulty of the grounding task, with referred objects that have only 2 distractors considered \textit{easy}, and objects that have up to 6 distractors are considered \textit{hard}. Second form of splitting considered view-dependency. Based on distinct word mining, utterances are classified as either view-dependent or view-independent.
% \vspace{-0.3cm}

\subsection{Results}
In Tab.~\ref{tab:results_referit} we report the performance of our proposed AsphaltNet on the Nr3D and Sr3D \textit{val}-sets and compare them to state-of-the-art methods. As seen, AsphaltNet is competitive with the existing work on both datasets, without relying on added supervision from the introduction of handcrafted detection prompts like BUTD-DETR (which account for $+4.2\%$ overall accuracy on Nr3D), or on boosting performance by averaging the multiple predictions from view augmented forward passes like MVT. 3DVista proposes a pretrained foundational model for 3D and text alignment, utilizing a rich and diverse dataset. We compare our results with 3DVista(scratch) which follows the same dataset and evaluation protocol as AsphaltNet, outperforming on both Nr3D (+1.4) and SR3D datasets.  
% Furthermore, it doesn't rely on  additional datasets as seen in approaches like 3DVista~\cite{3dvista} and Scanents~\cite{abdelreheem2022scanents}. 
% 3DVista proposes a pretrained transformer which is trained on a much larger and diverse dataset. Scanents relies on an augmented dataset which evaluated using privileged . We further do not compare to Vil3DRef ~\cite{chen2022language} as it utilizes an evaluation protocol that differs from the standard benchmark, rejecting many prompts in the test set. 
\subsection{Ablation Studies}

\begin{table}[t]

    \centering
    \tabcolsep=0.15cm
    \begin{tabular}{|ccc|c|cccc|}
        \hline
        $\mathcal{L}_o$ & TBA & $\mathcal{L}_{sp}$ & Overall & Easy & Hard & V-Dep & V-Ind \\
        \hline
        & & & 45.5 & 52.0 & 38.7 & 38.9 & 48.3 \\
        \ding{51} & & & 51.1 & 57.9 & 45.6 & 50.4 & 51.5 \\
        \ding{51} & \ding{51} & & 55.2 & 60.2 & 50.5 & 54.6 & 55.4 \\
        \ding{51} & \ding{51} & \ding{51} & \textbf{58.9} & \textbf{64.3} & \textbf{53.9} & \textbf{57.4} & \textbf{59.8} \\
        \hline
    \end{tabular}
        \caption{Ablation study showing the benefits on each introduced contribution. Starting with a baseline model using a top-down BAF module~\cite{unal2023ways}, we introduce our offset loss ($\mathcal{L}_o$), top-down bidirectional attentive fusion module (TBA), and the span loss ($\mathcal{L}_{sp}$). We report results on the NR3D \textit{val}-set [$\%$].}
\label{tab:ablation_modules}
\vspace{-0.4cm}

\end{table}

\begin{table}[t]
  \centering
  \tabcolsep=0.67cm
  \begin{tabular}{|l|c|c|}
    \hline
    Method & Overall & Failure Cases  \\
    \hline
    Baseline & 1.33 & 2.54\\ %% std 1.6294 , median, for the negative pred only: 
    $ + \mathcal{L}_o$ & 0.93 & 2.02 \\ %% std 1.5056, meadian 0.0, for the negative pred only : 
    \hline
  \end{tabular}
  \caption{Mean L2 distance of the predicted instance mask center to the referred instance mask center with and without the offset loss [m]. We report the distance across the validation-set (overall) along with only on failure cases where the model incorrectly predicts the referred object.}
  \label{tab:ablation_offsetloss}
\end{table}
\begin{table}[]
% \begin{minipage}{.48\textwidth}
    \centering
    \begin{tabular}{|c|c|}
    \hline
        Verbal Backbone & Overall Acc  \\
          \hline
          BERT\cite{devlin2018bert} & \textbf{58.9} \\
          ROBERTA \cite{liu2019roberta} & {58.8} \\
          MPNet\cite{song2020mpnet} & 56.1 \\
          \hline
    \end{tabular}
        \caption{Ablation Study showing the effect of changing the verbal backbone}

    \label{tab:verbal_backbone}
% \end{minipage}
\end{table}
\begin{table}[ht]
  \centering
  \begin{tabular}{|l|c|cc p{1cm}  p{1cm}|}
    \hline
     Method & Overall & Easy& Hard & V-Dep & V-Ind   \\
    \hline
    with $\mathcal{L}_{cls}$  & 56.3 & 62.6 & 50.6 & 55.5 & 56.9\\ %%%% need o change
    with $\mathcal{L}_{sp}$ &  \textbf{58.9} & \textbf{64.3} & \textbf{53.9} & \textbf{57.4} & \textbf{59.8} \\
    \hline
  \end{tabular}
  \caption{Comparison of the standard cross-entropy based sentence-level supervision ($\mathcal{L}_{cls}$) to our word-level span loss $\mathcal{L}_{sp}$.}
  \label{tab:SpanlossvsCls}
\end{table} 
% Easy acc :  0.6109762020398252
% Hard acc :  0.4793999104343932
% View Dependant :  0.5121344119477287
% \begin{table}[ht]
%   % \centering
%   % \tabcolsep=0.67cm
%   % \begin{tabular}{|l|c|c|c|}
%   %   \hline
%   %    Weight & 0.1 & 1 & 10  \\
%   %   \hline
%   %   $\mathcal{L}_o$ & 54.9 & \textbf{56.1} & 53.8 \\
%   %   $\mathcal{L}_{sp}$ & 55.8& \textbf{56.1}& 54.8  \\
%   %   % 0.1 & 54.9 \\ 
%   %   % 1 & 56.1 \\ 
%   %   % 10 & 53.8 \\ 
    
%   %   % Baseline & 1.33 \\ %% std 1.6294 , median, for the negative pred only: 
%   %   % + $\mathcal{L}_o$ & 0.96 & 2.19 \\ %% std 1.5056, meadian 0.0, for the negative pred only : 
%   %   \hline
%   % \end{tabular}
%   \caption{Hyperparameter study on the effect of varying weights of the offset and span losses on 3D visual grounding performance.}
%   \label{tab:ablation_lossweights}
% \end{table}

\begin{table}[t]
  \centering
  \tabcolsep=0.22cm
  \begin{tabular}{|l|c|c|cccc|}
    \cline{2-7}
     \multicolumn{1}{c|}{} & Weight & Acc & Easy & Hard & V-Dep & V-Ind \\
    \hline
    \multirow{3}{*}{\rotatebox[origin=c]{90}{$\mathcal{L}_o$}} & 0.1 & 55.7 & 61.1& 49.8 & 53.8& 56.8 \\
    
                           & 1 & \textbf{58.9} & 64.3 & 53.9&  57.4 & 59.8 \\
                           & 10 &  49.8 & 55.3 & 45.1 & 49.0 & 50.8 \\
    \hline
    \multirow{3}{*}{\rotatebox[origin=c]{90}{$\mathcal{L}_{sp}$}} & 0.1 & 55.8 & 61.1 & 50.9 & 54.6 & 56.4\\

                      & 1 &   \textbf{58.9} & 64.3 & 53.9&  57.4 & 59.8  \\
                         & 10 & 56.8 & 61.4 & 52.4 & 55.7 & 57.4 \\
    \hline

  \end{tabular}
    \caption{Hyperparameter study on the effect of varying weights of the offset and span losses on 3D visual grounding performance.}
  \label{tab:ablation_lossweights}
\vspace{-10px}\end{table}

\noindent \textbf{Effects of Network Components:}  In Tab.~\ref{tab:ablation_modules}, we perform an ablation study to investigate the effects of the different components of our proposed AsphaltNet. We set up a baseline model using a top-down attentive fusion module from ConcreteNet~\cite{unal2023ways}. We then systematically introduce our offset loss ($\mathcal{L}_o$), the top-down bidirectional attention block (TBA), and finally the span loss ($\mathcal{L}_{sp})$. As observed, the introduction of the offset loss drastically improves \textit{hard} grounding performance by $+6.9\%$. As mentioned above, the \textit{hard} category consists of referred objects with more than 3 same class distractors within the same scene, which illustrates the effectiveness of the offset loss in candidate separation as a byproduct of grounding-aware feature routing.

Following the offset loss, we introduce the bidirectional attention layers into the top-down fusion module, allowing the word tokens to get updated based on instance-level features. As seen, the overall accuracy further improves by $+4.1\%$. Unlike 2D images, since 3D scenes don't inherently possess an orientation, view-dependent prompts in Nr3D are often constructed through object references, e.g. ``If you looking at the whiteboard it's the chair on the back left." Such cases can be ill-posed when language tokens are unable to obtain information from the 3D scene. The inclusion of the top-down bidirectional attentive fusion block especially aids the \textit{view-dependent} accuracy, promoting scene-aware word encoding.

Finally, we introduce supervision to the word tokens via the span loss ($\mathcal{L}_{sp}$). As seen in Tab.~\ref{tab:ablation_modules}, the loss shows benefits across all categories, with the overall accuracy increasing by $+3.7$. With all three components, AsphaltNet reaches $58.9\%$ overall accuracy on Nr3D dataset.

\noindent \textbf{Impact of $\mathcal{L}_o$ on Localization}: In Tab.~\ref{tab:ablation_offsetloss} we conduct an ablation study to investigate the effect of the offset loss on referral-based object localization. Given the baseline model, we compute the average L2 distance of each model prediction to the referred target object with and without the inclusion of the offset loss during training.
Furthermore, we report the same metric over only failure cases to eliminate the bias stemming from the increased number of accuracy (i.e. 0 distance) samples. As seen, in both overall and failure cases, the inclusion of the offset loss reduces the average distance of the predicted object to the target object. In other words, the offset loss enables better localization within the network, promoting predictions nearer to the target object even during failure cases.

Additionally in Fig.~\ref{fig:offset_loss} we showcase a qualitative example of the evolution of the offset predictions within AsphaltNet. Specifically on each layer, for each instance, we obtain the predicted target object centroid and fit a 2D kernel density function~\cite{wand1994kernel} to the obtained set. As expected, the distribution of the centroid estimations converges on the referred object with increased precision, i.e. decreased uncertainty.
\vspace{3px}

\noindent \textbf{Impact of the Verbal Backbone}: In Tab.~\ref{tab:verbal_backbone} we provide an ablation study with different verbal encoding choices. We show that our model is robust against this choice, with BERT~\cite{devlin2018bert} and RoBERTa~\cite{liu2019roberta} achieving $58.9\%$ and $58.8\%$ overall accuracy. We observe a decrease in performance when employing MPNet which has seen previous success in 3D visual grounding models~\cite{unal2023ways}. We account this drop in performance towards the permuted language modeling of MPNet, which fails to capture the relational contexts in more structured ReferIt3D prompts adequately.

\noindent \textbf{Auxiliary Supervision on Language:}In Tab.~\ref{tab:SpanlossvsCls} we compare our proposed span loss $\mathcal{L}_{sp}$ to the standard approach of constructing an auxiliary classification problem using the language tokens and supervising it via a cross-entropy $\mathcal{L}_{cls}$~\cite{chen2020scanrefer, zhao2021_3DVG_Transformer}. As seen, providing a higher granularity in the language supervision aids the model performance, boasting an additional $+2.6\%$ overall accuracy over the cross-entropy formulation, with the majority of the benefits coming from the \textit{hard} category where utilization of language becomes critical to distinguish the referred object between a large number of semantically similar candidates.

\noindent \textbf{Hyperparameter Study:} Having introduced two additional losses to aid 3D referral-based object identification, we explore the effect of loss weights on the models performance. In Tab.~\ref{tab:ablation_lossweights}, we train AsphaltNet with varying weights on either the offset loss $\mathcal{L}_o$ or the span loss $\mathcal{L}_{sp}$. While the model is quite robust to the weight changes of the losses -- showing comparable performance to the second-best performing method despite an order of magnitude change in any loss weight -- the best configuration is obtained by setting each weight to 1.

\section{Conclusion}

In this work we tackle the problem of 3D visual grounding, where the goal is to identify the correct instance based on a natural language prompt. We note that while continual efforts are put into designing better architectures, the exploration of losses remains stagnant.  This work proposes two auxiliary losses to improve grounding performance and increase training robustness. Firstly we construct an offset loss to aid localization where each object aims to predict its offset to the referred target. Next, we extend our baseline method with bidirectional attention to allow information flow from the 3D scene to the language branch. This allows us to construct a span loss, providing highly granular word-level supervision to the verbal encoding. Combining all three components, we construct AsphaltNet, a novel 3D visual grounding approach with impressive performance on ReferIt3D without relying on any additional data.

\noindent\textbf{Acknowledgments:} This work is funded by Toyota Motor Europe via the research project TRACE-Z\"urich.

{\small
\bibliographystyle{ieee_fullname}
\bibliography{egbib}
}

\clearpage
\clearpage

\appendix

\section{Dataset} We evaluate our method on the Nr3D and Sr3D datasets~\cite{achlioptas2020referit_3d} that provide human-annotated and synthetically generated utterances to identify instances on ScanNet~\cite{dai2017scannet}.
Specifically, Nr3D consists of 41,503 natural language queries and Sr3D consists of 83,572 descriptions over 707 unique indoor scenes referring to 76 object classes. For evaluation, the dataset is split in two ways. First split considers the difficulty of the grounding task, with referred objects that have only 2 distractors considered \textit{easy}, and objects that have up to 6 distractors are considered \textit{hard}. Second form of splitting considered view-dependency. Based on distinct word mining, utterances are classified as either view-dependent or view-independent.
% \vspace{-0.3cm}
\section{Results and Evaluations}
\subsection{Multiple Objects} 
\begin{figure}
    \centering
    \includegraphics[width=\columnwidth]{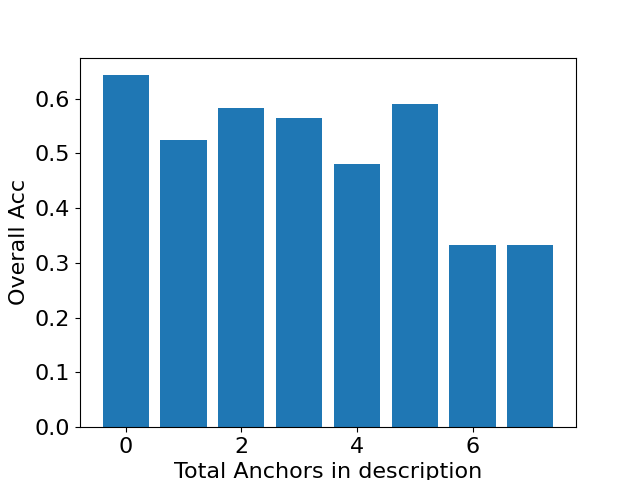}
    \caption{Plot visualizing the overall accuracy against the length of the natural language description \#tokens.}
    \label{fig:anchor_acc}
\end{figure}

Description referring to multiple objects is a compelling research problem that the ReferIt3D benchmark doesn't address. We carry out further analysis of our approach using the ScanEnts dataset ~\cite{abdelreheem2024scanents3d}. to evaluate the network performance with the number of anchors in the utterance \ref{fig:anchor_acc}. A drop in performance with more objects in the utterance is observed. 
\begin{table}[]
    \centering
 \begin{tabular}{|c|c|}
        \hline
         Method & Overall  \\
         \hline
        Top-down& 58.9\\
        Bottom-up & 56.3\\
        \hline
    \end{tabular}
    \caption{Comparison of top-down and bottom-up attention masking
for the verbo-visual fusion module}
    \label{tab:topdown}
\end{table}
\section{Top-Down Masking} To explicitly limit information routing within local neighborhoods, ConcreteNet~\cite{unal2023ways} employs a masking operation that hinders self-attention between spatially distant instance candidates. This masking operation is implemented in a bottom-up manner where the masking sphere grows as the layers increase.
To be able to utilize the offset loss introduced in the main manuscript at various stages of the bidirectional attentive fusion module, we instead use a top-down approach, allowing all instances to be aware of one another even at the shallowest levels. In Tab.~\ref{tab:topdown}, we ablate this masking approach and show that a top-down method works better than a bottom-up one.

\subsection{Out of Distribution} We address the out-of-distribution ability of AsphaltNet. In table~\ref{tab:OOD-Scanrefer} we evaluate our network on the scanrefer dataset~\cite{chen2020scanrefer} with GT boxes provided in the input with network weights corresponding to NR3D dataset. The network trained without any of the contribution of AsphaltNet is used as baseline for the comparisons. We observe decent performance on the scanrefer dataset given the vast difference in the language domain of ReferIt3D and scanrefer datasets showcasing the OOD robustness of AsphaltNet. We further investigate the synthetic to real transferability of the network. AsphaltNet trained on SR3D dataset is tested on the NR3D dataset to test for synthetic - real domain adaptability. AsphanltNet yields relatively lower performance , table ~\ref{tab:OOD-NR3D}, when evaluated on NR3D dataset. This can attributed to a lack of diversity in the synthetic dataset hence unable to transfer to the natural language settings.  

\begin{table}[]
    
\begin{minipage}{.23\textwidth}

    \centering
    \begin{tabular}{|c|c|}
        \hline
        Method & Accuracy  \\
         \hline
         Baseline & 43.89 \\ 
         AsphaltNet & 47.99 \\
         % \\ Vil3dref & 53.48
         \hline
    \end{tabular}
    \caption{Evaluation of OOD performance on ScanRefer}
    \label{tab:OOD-Scanrefer}
\end{minipage}
\hfill
\begin{minipage}{.23\textwidth}
    \centering
    \begin{tabular}{|c|c|}
        \hline
        Method & Accuracy  \\
         \hline
         Baseline & 34.87 \\ 
         AsphaltNet & 38.53 \\
         % \\ Vil3dref & xx
         \hline
    \end{tabular}
    \caption{Evaluation of OOD performance on NR3D}
    \label{tab:OOD-NR3D}
\end{minipage}
\end{table}

\begin{table*}[h]

  \centering
  \tabcolsep=0.3cm
  \begin{tabular}{|c|l|c|cccc|}
  \cline{2-7}
   \multicolumn{1}{c|}{}  & Network & Overall & Easy & Hard & View-Dep & View-Ind  \\
    % \hline
    \hline
    % \multirow{10}{*}{\rotatebox[origin=c]{90}{Nr3D}}
    % \multirow{4}{*}{\rotatebox[origin=c]{0}{NR3D}} & MVT (Single view) & 51.6 & 58.4 & 45.0 & 50.7& 52.0\\ 
    \multirow{2}{*}{\rotatebox[origin=c]{0}{NR3D}}&AsphaltNet & 58.9 & 64.3 & 53.9 & 57.4 & 59.8  \\ 
    % \cline{2-7}
    % \hline
    % &MVT (Multi view) & 55.1 & 61.3& 49.1 & 54.3& 55.4\\ 
    &AsphaltNet + MVE & \textbf{59.3} & 64.9 & 54.2 & 57.8 & 59.9 \\
    \hline 
    
     % \multirow{2}{*}{\rotatebox[origin=c]{0}{SR3D}}  & MVT (Multi view) & 64.5  & 66.9  & 58.8  & 58.4  & 64.7\\ 
     
    \multirow{2}{*}{\rotatebox[origin=c]{0}{SR3D}} & AsphaltNet & 69.7 & 71.9 & 64.5 & 67.2 & 70.0\\

    & AsphaltNet + MVE & \textbf{72.7} & 74.6 & 68.2 & 70.4 & 73.0\\
    % Asphaltnet & 0.96 & 2.19 \\ %% std 1.5056, meadian 0.0, for the negative pred only : 
    \hline
  \end{tabular}
  \caption{Evaluation of  AsphaltNet with multi-view ensembling (MVE)} 
  \label{tab:MVT_comparison}
\end{table*}

\subsection{Multi-View Ensembling}
MVT~\cite{huang2022multiview} proposes a multi-view feature aggregation approach that aggregates visual features from different viewpoints during both training and testing. While this acts as a form of regularization during training with the goal of smoothing the gradients, during test time, the multi-view aggregations act as a form of test-time-augmentation (TTA). 
AsphaltNet's performance can be enhanced by utilizing TTA by employing multiple forward passes through affine-transformed point clouds. Specifically, we employ multi-view ensembling (MVE) following ConcreteNet~\cite{unal2023ways} where we infer a scene $N$ times, each with the point cloud rotated by an angle $r \in [0,2\pi]$. The final prediction is determined through majority voting across the $N$ inferences (with $N=9$). As seen in table ~\ref{tab:MVT_comparison}, MVE further boosts the performance of AsphaltNet for both NR3D and SR3D splits of ReferIt3D.

\begin{figure}[h]
    \centering
    \includegraphics[width=\columnwidth]{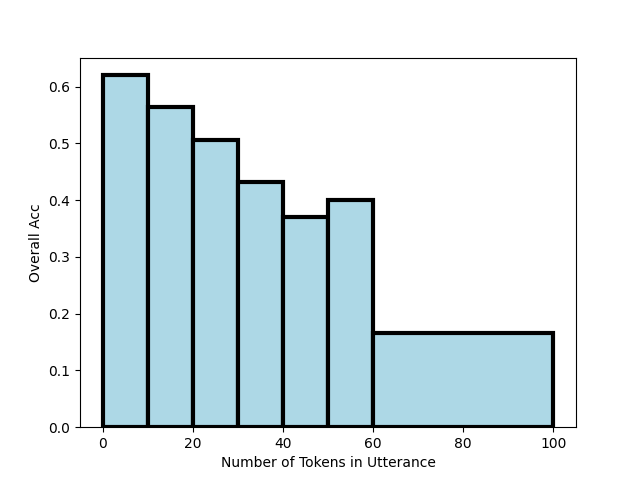}
    \caption{Plot visualizing the overall accuracy against the length of the natural language description \#tokens.}
    \label{fig:acctoken}
\end{figure}
\begin{figure}[h]
    \centering
    \includegraphics[width=\columnwidth]{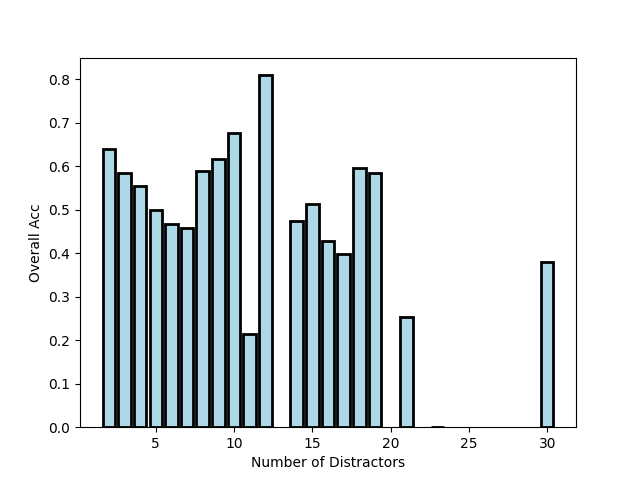}
    \caption{ Plot visualizing the overall accuracy against the total number of anchors in the description.}
    \label{fig:accdistractor}
\end{figure}
\section{Limitation and Discussion}

While AsphaltNet achieves state-of-the-art results on both ReferIt3D~\cite{achlioptas2020referit_3d} benchmarks, we still observe an important limitation when analyzing the predictions. In Fig.~\ref{fig:acctoken} we show the accuracy of our model on the Nr3D \textit{val}-set against the number of tokens in the natural language prompt. As seen, we observe a linear drop in performance as the number of tokens increases. We speculate this is due to the increased difficulty of the span prediction task that limits our method's generalization ability for longer text descriptions.

Furthermore, in Fig.~\ref{fig:accdistractor}, we show that while our method's performance weakens as the number of tokens increases (verbal input), the same behavior is not observed when investigating the visual input changes, i.e. as the number of same-class distractors increases. This illustrates the benefits of the offset loss, where the increased robustness towards spatial localization reduces the negative effects of increased same-class instances within the scene.

\end{document}